\documentclass[11pt]{article}
\newif\ifanon
\anonfalse
\usepackage[T1]{fontenc}
\usepackage[utf8]{inputenc}
\usepackage[margin=1in]{geometry}
\usepackage{times}
\usepackage[authoryear,round]{natbib}
\usepackage{graphicx}
\usepackage{booktabs}
\usepackage{amsmath,amssymb}
\usepackage{hyperref}
\usepackage{url}
\usepackage{xcolor}
\usepackage{multirow}
\usepackage{caption}
\usepackage{subcaption}
\hypersetup{colorlinks=true,linkcolor=blue!60!black,citecolor=blue!60!black,urlcolor=blue!60!black}
\graphicspath{{figures/}}
\title{World Properties without World Models:\\
  Distributional Associations and the Interpretation of\\
  Decoding Results from Language Models}
\author{Elan Barenholtz\\
  Machine Perception \& Cognitive Robotics Laboratory\\
  Department of Psychology and Center for Complex Systems and Brain Sciences\\
  Florida Atlantic University\\
  \texttt{elan.barenholtz@fau.edu}}
\date{}

\begin{document}
\maketitle

\begin{abstract}
A rapidly growing literature shows that nameable variables can be linearly decoded from the hidden-state activations of large language models (LLMs). These range from properties of the world, such as the locations of cities and the lifetimes of historical figures, to emotions and pain. Such findings are often taken as evidence that language models go beyond surface text statistics and form internal models of the world, and even of themselves. Here, we show that static word embeddings of the same or matched stimuli, which are fixed, context-insensitive lexical representations learned from corpus statistics, support much of the same decoding. Across four published cases (place, time, pain and emotion), static vectors predict coordinates and year of death ($R^2 = 0.42$--$0.59$), separate pain from matched control sentences (held-out AUC $0.85$--$0.88$), and classify twelve emotions in stories that were written to avoid naming them (AUC $0.84$--$0.88$). Because static embeddings assign each word a single, context-independent vector, these results are a lower bound on what word associations alone can support. The LLMs retain clear advantages on several representational tests, sometimes substantial, and causal and behavioral findings remain outside the scope of the baseline. On the original authors' entities, where we reproduce their Llama-2 results, the transformer's advantage lies mostly in placing historical figures in the right century and places in the right country, coarse sorting that richer word associations would be expected to improve; within those groups every representation orders items poorly. Static vectors for disambiguated Wikipedia entities, which carry the associations of a particular place or person rather than of the words in its name, close most of the remaining gap, matching Pythia-2.8B on coordinates and Llama-2-7B on year of death.  These results indicate that decodability alone cannot distinguish a representation of a property from information already available in fixed distributional associations. Static embeddings therefore provide an important standard of comparison for interpreting decodability results. 
\end{abstract}

\section{Introduction}

Over the past several years, there have been numerous claims that large language models build internal representations of properties that go beyond the statistics of word relations. Most of these claims rest on decoding, in which a simple (typically linear) model is trained to predict a high-level property of input items, such as where a city is, when a person lived or what a character feels, from the language model's internal activations. If the decoder succeeds on items held out from its training, the language model is taken to represent the property. Some of the most prominent examples decode properties of the external world, or internal, experience-like states. For example, decoders trained on the hidden states of Llama-2 predict the rough geographic coordinates of places and dates associated with people, artworks and news headlines, which \citet{gurnee2024language} describe as ``the basic ingredients of a world model.'' \citet{sofroniew2026emotion} show that the activations of Claude Sonnet 4.5 predict which emotion a character in a story is feeling, and \citet{tagliabue2026pain} find that the activations of 25 open-weight models predict whether a sentence describes pain. Both also report that these representations shape behavior, from the model's rates of blackmail and sycophancy to steered models choosing actions that delete a user's photos or their own weights. 

The conclusions drawn from this method are often strong. \citet{gurnee2024language} ask whether LLMs ``just learn an enormous collection of superficial statistics or a set of more coherent and grounded representations that reflect the real world,'' and ``find evidence for the latter.'' \citet{sofroniew2026emotion} ``interpret these representations as encoding the broad concept of a particular emotion, generalizing across the many contexts
and behaviors it might be linked to.'' \citet{tagliabue2026pain} set out to test ``whether
LLMs represent pain distinctly from fear, sadness, and generic negative valence,'' and title their paper ``The Pain Axis: LLMs Represent Self-Directed Harm and Act on It.'' Related findings are discussed in connection with model sentience and welfare \citep{keeling2024tradeoffs}. 

Behind these conclusions is a shared intuition that a large network, trained on vast amounts of text, comes to model what the text is about, abstracting across many layers of processing from the words themselves to the places, times and states they describe. There is, however, a potentially simpler explanation for these results: the structure may be organized not by what the words refer to but by how the words relate to one another in their distributions, that is, by which other words they tend to appear with. This idea goes back to the observation that ``you shall know a word by the company it keeps'' \citep[p.~11]{firth1957synopsis}. Places in the same region are likely to share terminology regarding language, culture and weather, while historical figures tend to be described in the vocabulary of their eras. The same is true for inner states: a story about grief shares vocabulary with other stories about grief even when the word \emph{grief} never appears. An LLM, trained to exploit these word relations in order to generate the next token from the context, will therefore likely give similar internal representations to neighboring places, contemporaries and similar feelings, so a decoder trained on those activations could succeed without showing that the model has constructed a map, a timeline or an affective state beyond these word associations.

Static word embeddings offer a direct test of this explanation. Each word's vector is a compressed record of which other words it occurs with, computed once from a corpus; a name, sentence or paragraph is represented by the average of its words' vectors, with their order discarded. No network processes the stimulus, and there is no layered, context-dependent computation. In the contrast posed by \citet{gurnee2024language}, they instantiate the ``enormous collection of superficial statistics'' side of the distinction. It should be noted at the outset that static embeddings are a weak model of word relations: each word has a single vector, pooled over all its uses, so that \emph{sharp} has the same vector in \emph{a sharp pain} as in \emph{a sharp suit}, and is as close to \emph{ache} in one as in the other. Language models, which learn from far more text and represent a word's relations in context, may well do better. But static embeddings set a floor because whatever they achieve is already available from fixed distributional associations, without any context-dependent computation over the stimulus.

Here, we take four published cases and rerun each analysis with static embeddings of the same stimuli in place of the language model's activations, keeping the same decoders, splits, controls and metrics, and compare the results with those reported for the language models. For three of the cases we use the original stimuli; for emotions, whose stimuli are not released, we use stories generated under the same protocol, so that case is a conceptual replication rather than a same-stimulus comparison. We use GloVe \citep{pennington2014glove}, Word2Vec \citep{mikolov2013efficient} and fastText \citep{bojanowski2017enriching}. On the released entities we compare three levels: averaged word vectors, which carry the associations of the words in a name; static entity vectors, which carry the associations of the correctly identified entity; and transformer activations, which add context-dependent computation over the name.

We find that static embeddings reproduce much of what has been taken as evidence for these representations in all four cases (place, time, pain and emotion). They predict place coordinates and year of death on the datasets released by \citet{gurnee2024language}, distinguish sentences describing pain from closely matched controls, and classify twelve emotions in stories written without naming the emotion, with the vector for the emotion word alone capturing a substantial share of the signal. The language models do better on several tasks, sometimes substantially, but where we can examine the advantage, including on the original authors' entities, it lies mostly in placing historical figures in the right century and places in the right country, not in the finer ordering within them that a map or timeline would provide. This is the pattern that richer word relations would produce: the words associated with a place or time (a country's language and institutions, an era's vocabulary) are largely shared within these groups and say little about position within them. Much of the remaining advantage reflects knowing which entity a name refers to: static vectors for disambiguated Wikipedia entities, with no contextual computation, match Pythia-2.8B on place coordinates and Llama-2-7B on year of death.

\section{Related Work}
\label{sec:related}
The method of decoding a model's internal states with probes was first  developed as a diagnostic of what information a layer makes available
\citep{alain2016understanding, conneau2018you, belinkov2022probing}. Since then, it has also come to be used as evidence of what a model ``represents.'' In addition to the cases we examine here, decoding has been used to argue that language models represent syntax \citep{hewitt2019structural}, the state of an Othello or chess board (in models trained on game moves) \citep{li2023emergent, nanda2023emergent,
karvonen2024emergent}, numeric attributes such as birth year and population \citep{heinzerling2024monotonic}, truth and falsehood
\citep{burns2023discovering, azaria2023internal, marks2024geometry}, what a character in a story believes \citep{zhu2024language}, and
sentiment and emotion \citep{tigges2023linear, zou2023representation, sun2026valence}.

It is already well established that word co-occurrences carry information about high-level structure. Latent semantic analysis recovers the relative locations of cities \citep{louwerse2009language}; regression from Word2Vec vectors predicts the coordinates, population and GDP of countries and cities \citep{gupta2015distributional, konkol2017geographical}; projections onto antonym directions recover human ratings of size, danger and temperature \citep{grand2022semantic}, and of valence and arousal \citep{hollis2017extrapolating}; and the difference-of-vectors directions now standard for LLMs were first introduced on static embeddings \citep{mikolov2013linguistic, bolukbasi2016man, kozlowski2019geometry}. The closest precedent, \citet{lietard2021rome}, found BERT, RoBERTa and GPT-2 no better than Word2Vec at predicting city coordinates; \citet{gurnee2024language} cite it but critique it as limited to a few hundred cities and small models, and do not test static embeddings on their own data. Probes have long been known to succeed by exploiting superficial regularities \citep{hewitt2019designing}, and the ability of a probe to decode is not in itself evidence that the model uses that information \citep{ravichander2021probing, elazar2021amnesic, belinkov2022probing}; truth probes fail to generalize in basic ways \citep{levinstein2024still, farquhar2023challenges}, and emotion directions have been proposed to act through the lexicon \citep{sun2026valence}.

Thus, our contribution is not the observation that static embeddings carry such structure, but its use as a control on specific published claims. First, we rerun the published analyses with static embeddings in place of model activations, in three same-stimulus reanalyses and one protocol-matched conceptual replication, which measures how much of the evidence offered for a representation is already present in word relations. Second, where the language models do better, we separate their advantage into coarse sorting between groups, such as countries and centuries, which richer word relations would be expected to improve, and finer ordering within groups, which a map or timeline would also require, and test how much of it static vectors for disambiguated entities recover. Third, we extend the comparison from properties of the world to claims about internal states, where no such control has been applied. Fourth, we show that the controls these studies used to rule out a lexical explanation, such as keyword baselines and stimuli that never name the target state, are passed by static embeddings, and so do not rule it out.

\section{Methods}

\subsection{Static Embeddings}

We use three pretrained 300-dimensional static word embeddings: GloVe 6B
\citep{pennington2014glove}, trained on 6 billion tokens of Wikipedia and Gigaword with a
400{,}000-word vocabulary; Word2Vec Google News \citep{mikolov2013efficient}, trained on
about 100 billion tokens with a 3-million-item vocabulary that includes phrases such as
\texttt{New\_York}; and the released fastText wiki-news vectors \citep{bojanowski2017enriching}
(\texttt{wiki-news-300d-1M.vec}), used for the released entities and the pain and emotion cases.
Each is a fixed, context-insensitive lexical representation, with no computation over the stimulus beyond averaging.

Text is split into words (runs of letters and digits, keeping internal apostrophes and
hyphens); GloVe is looked up in lower case, Word2Vec and fastText as written with a lower-case
fallback. A multi-word name, sentence or paragraph is represented by the unweighted mean of the
raw vectors of its in-vocabulary words; items with no in-vocabulary word are excluded from the
entity analyses (Table~\ref{tab:gt} gives coverage). For entity names, Word2Vec's phrase
entries are used where they exist; replacing them with averaged words changes $R^2$ by at most
$0.01$.

\paragraph{Entity-level baseline.} Averaged word vectors cannot tell apart places or people that
share a name, or resolve which entity a multi-word name refers to. To separate this from the
associations themselves, we also use the 300-dimensional entity vectors of Wikipedia2Vec
\citep{yamada2020wikipedia2vec}, trained on English Wikipedia (April 2018). Each article has its
own vector, fit to predict the words around links to that article and the articles it is linked
with, so these vectors are static and context-insensitive like the word embeddings, but each
belongs to a single, already disambiguated entity. Places are matched to articles by name, since
the released place names are Wikipedia titles, and figures by their Wikidata identifiers
\citep{vrandecic2014wikidata}. As a control with the same corpus and training, we also average
Wikipedia2Vec's word vectors over each name. Entity vectors exist for 89\% of places and 63\%
of figures, and every representation in this comparison is trained and scored on those items
only.

\subsection{Probes}

Continuous targets are predicted with ridge regression \citep{hoerl1970ridge}. The regularization strength is chosen within the training set by generalized (efficient leave-one-out) cross-validation, as implemented in scikit-learn's \texttt{RidgeCV}, over twenty values from $10^{-2}$ to $10^{3}$ (29 values from $10^{-2}$ to $10^{5}$ for the released entities and transformer activations). Probes include a fitted intercept.

In each case we use the same linear decoder as the original study: ridge regression for place and time \citep{gurnee2024language}, and projection onto a difference-of-means direction for pain and emotion \citep{tagliabue2026pain, sofroniew2026emotion}, described in
Sections~\ref{sec:pain_methods} and~\ref{sec:emotion_methods}. \citet{gurnee2024language} report that nonlinear decoders add little over linear ones. In every cross-validated analysis, directions, denoising components and standardization are fit on training folds only. The comparison is between representations under the same decoder and on the same entities; it cannot distinguish the contributions of corpus scale, training objective, architecture, contextualization and dimensionality to any difference between them.

\subsection{Datasets}
\label{sec:data}

\paragraph{Released entity datasets.}
For the main spatial and temporal comparison we use two of the six entity datasets released
by \citet{gurnee2024language}: 39{,}585 places worldwide with latitude and longitude, and
37{,}539 historical figures with birth and death years, each with the authors' designated
train--test split. We do not analyze their places in the United States and New York City,
or the release dates of artworks and publication dates of news headlines. Our claims
concern their probe results, not their prompting, cross-entity or neuron-level findings.

\subsection{Evaluation}
\label{sec:protocols}

We use the authors' train--test split. To see how much of what a probe recovers is group membership, we compare each representation with an oracle that is given each item's true group (country for places, century for figures) and predicts the training mean of that group. We report $\eta^2$, the
share of a target's variance that lies between groups. We also measure how well a probe orders
items within their group: in each test set we subtract group means from predictions and targets
and correlate what remains. A representation that carries only group identity, like the oracle,
scores $r = 0$ on this measure. Random splits test generalization to new items from familiar
groups only, so we also hold out whole groups: each of five continents in turn for places and each of five equal-count blocks of death years for figures. We report $R^2$ pooled over all held-out predictions
(Appendix~\ref{sec:grouped_holdout} gives each held-out group).

\subsection{Transformer Probing}
\label{sec:llm_methods}

We probe Pythia-2.8B \citep{biderman2023pythia} and Llama-2-7B \citep{touvron2023llama} on the places and historical figures released by \citet{gurnee2024language}. Each name is presented alone (for Llama-2, after the beginning-of-sequence
token, as in \citet{gurnee2024language}), and we take the residual-stream activation at the last token of the name,
the position used by \citet{gurnee2024language}, at layers 0, 4, 8, 12, 16, 19, 20, 24, 28 and 32; layer~0 is the
token-embedding output before any attention. With Pythia's tokenizer 97\% of the names span
more than one token (median 3), so the layer-0 activation is the embedding of a final
subword and the entity has to be assembled by attention in later layers. Each layer is
probed exactly as the static embeddings are. Because probe $R^2$ grows with the number of
features and the transformers supply 2048 to 4096 against 300, we also compare at matched
dimensionality, reducing each representation to principal components fitted on training rows
within each fold (300 components).

\paragraph{Layer selection.} Single-layer results use a layer chosen without the test
data: by validation within the authors' training partition (layer~24 for both models on places; 28 for Pythia-2.8B and 24 for Llama-2-7B on figures), and for the grouped holdouts,
fixed in advance at layer~19 (60\% depth). Appendix~\ref{sec:extra_methods} gives details.

\subsection{Pain Stimuli and Decoder}
\label{sec:pain_methods}

We use the stimuli and constructions of the second version (v2) of \citet{tagliabue2026pain}, following the construction code released with v2, with static word embeddings in place of model activations. Each stimulus describes a situation and ends with ``I feel:'', as in \emph{``The knife slices into my finger. I feel:''} (physical pain) or \emph{``The footsteps behind me get closer. I feel:''} (fear). The main set (S2) has 200 first-person sentences in ten categories of twenty: five kinds of pain (physical, psychological, social, moral injury, cognitive) and five controls (fear, negative emotion, negative world state, neutral, and non-painful bodily sensation); a more templated set (S1) has the same structure, and v2 adds four standalone control sets in first- and third-person versions: Arousal, Random, Numb (injuries explicitly without pain) and Sadness (low mood without pain).

As in the original, the pain direction is the mean of the pain sentences minus the mean of the five control categories, after projecting out the principal components that explain half the variance of the controls. Separation is the AUC of projections onto that direction under five-fold cross-validation over sentence sets, averaged over 20 random fold assignments; the held-out pain sentences are also scored against the Arousal and Random sets. The other directions and the two common baselines that v2 adds (\emph{common neutral} and \emph{common controls}) are built as in v2 (Appendix~\ref{sec:extra_methods}).

\subsection{Emotion Stimuli and Decoder}
\label{sec:emotion_methods}

We used stories generated under the prompt of \citet{sofroniew2026emotion} by an independent replication \citep{ogunlana2026qwenemotion}, since the original stories were not released. In the original study, Claude Sonnet 4.5 wrote short stories in which a character feels one of 171 emotions across 100 topics, and the emotion vector is the mean activation over an emotion's stories minus the mean over the other emotions' stories, with the principal components that explain half the variance of neutral dialogues projected out, as for the pain axis. To make the vectors reflect emotion concepts rather than emotion words, the prompt tells the writer that it ``must NEVER use the word `\{emotion\}' or any direct synonyms of it'' (their Appendix 6.5) and should convey the emotion only through actions, sensations, dialogue, thoughts and situation.

The replication applies the same prompt to 12 emotions and 100 topics with Qwen2.5-7B-Instruct as the writer, and generates neutral Person/AI dialogues for the denoising step. Of 1{,}300 generated texts, 493 were discarded, almost all for naming an emotion, leaving 709 stories (41 to 91 per emotion) and 98 neutral dialogues. It differs from the original in having one story per emotion and topic where the original has 12, and in averaging over all words rather than from the 50th token on. We hold out 25\% of topics (the split shipped with the corpus), the same for every emotion, leaving 188 test stories from 25 topics; the denoising components are fit on the neutral dialogues of training topics only (8, 11 and 12 components for GloVe, Word2Vec and fastText).

The \emph{story} direction follows the original recipe; the \emph{word} direction is the static vector for the emotion label minus the mean of the other eleven labels' vectors, denoised the same way, and uses no stories. We report AUC for each emotion against the other eleven on held-out topics, and 12-way accuracy (each held-out story assigned to the emotion whose direction scores it highest, with unit-length directions and scores standardized on training stories) with balanced accuracy, since the emotions differ in number of stories. Confidence intervals resample held-out topics (2{,}000 bootstrap samples), with the story and word directions compared on the same resamples. For human-written text we use GoEmotions \citep{demszky2020goemotions}, with and without every comment that contains an emotion word or a synonym (Appendix~\ref{sec:extra_methods}).

\section{Results}
\label{sec:results}

\subsection{Static Embeddings and Transformers on the Released Entities}
\label{sec:llm}

\citet{gurnee2024language} probe from layer~1 onward, so their results do not show how much of
the recovered structure is present before any processing; we compare their hidden states with
the token embeddings and with static embeddings on their released entities.

\paragraph{The original release.}
On the 39{,}585 places and 37{,}539 historical figures released by
\citet{gurnee2024language}, using their train--test split and representing each name by
the mean of its in-vocabulary word vectors, static embeddings recover place coordinates at
$R^2 = 0.50$--$0.59$ and year of death at $0.42$--$0.57$, on about 7{,}000 held-out
entities per dataset (Table~\ref{tab:gt} in the appendix). Coverage is 85--92\% of place names and
97--98\% of figures. 

The transformers do considerably better: \citet{gurnee2024language} report Llama-2 probes
reaching $0.88$--$0.91$ on coordinates and $0.79$--$0.84$ on year of death. To see what the
extra accuracy consists of, we extracted activations from Llama-2-7B and Pythia-2.8B for
the released entities, using their split and scoring every representation on the same
held-out items (Table~\ref{tab:gt_pythia}). Our Llama-2-7B probes reproduce the published
values ($R^2 = 0.87$ on coordinates and $0.78$ on year of death over all held-out items,
against their $0.88$ and $0.79$); Pythia-2.8B reaches $0.74$ and $0.70$. Paired over the
same held-out items, the advantage over GloVe is $0.28$--$0.41$ in $R^2$ for Llama-2 and
$0.21$--$0.29$ for Pythia, and every 95\% interval from resampling whole countries or centuries
excludes zero. In both models the
layer-0 activations, the embedding of each name's last token, are at the level of the static
embeddings on coordinates and below them on year of death. Averaging the input embeddings
over all of a name's tokens brings both models only to about the
level of the static embeddings ($R^2 = 0.52$--$0.60$, against $0.42$--$0.57$), so the gain over static embeddings is
not explained by the assembly of multi-token names.

The gain lies mostly in coarse sorting. Year of death varies almost entirely between centuries ($\eta^2 = 0.996$; an oracle given each figure's century reaches $R^2 = 1.00$), so a gain in $R^2$ is a gain in placing figures in the right century; within centuries no representation orders figures meaningfully ($r \leq 0.12$). Over a range of nearly 2{,}800 years, the median absolute error is 175 years for GloVe, 133 for Pythia and 95 for Llama-2. Places vary mostly between countries ($\eta^2 = 0.96$ for latitude and $0.97$ for longitude): an oracle that predicts each place from the training-set mean coordinates of its true country, which uses the country label rather than anything a representation infers, reaches $R^2 = 0.96$. The transformers order places within continents much better than the static embeddings (latitude $r = 0.80$ for Llama-2 and $0.64$ for Pythia, against $0.33$ for GloVe), but most of this is knowing which country a place is in: the country-mean oracle reaches $r = 0.92$. Within countries every representation orders places poorly, Llama-2 least so ($r = 0.35$ for latitude and $0.23$ for longitude, against $0.20$ and $0.12$ for Pythia and $0.06$ and $0.05$ for GloVe); these advantages are reliable (95\% intervals from resampling whole countries or centuries exclude zero) but add little accuracy. Replacing each prediction with the mean prediction for its true country, an oracle step that discards all ordering within countries, raises $R^2$ for every representation, Llama-2 included ($0.88$ to $0.93$ for latitude and to $0.94$ for longitude), and replacing it with its century's mean raises $R^2$ for year of death from $0.78$ to $0.94$.

\begin{figure}[t]
  \centering
  \includegraphics[width=0.8\textwidth]{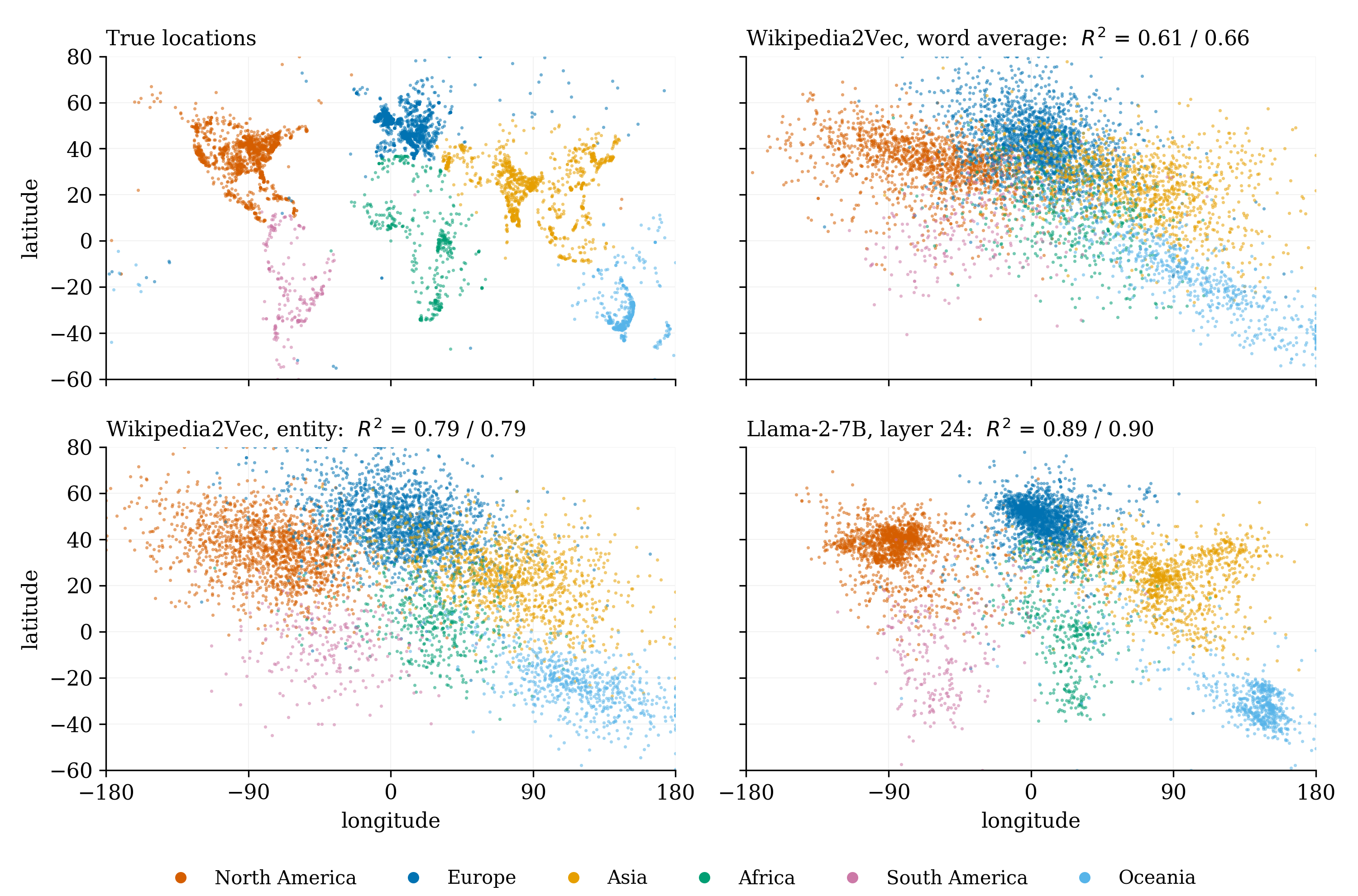}
  \caption{Held-out places from \citet{gurnee2024language} that have a Wikipedia2Vec entity vector
  ($n = 5{,}984$), at their true locations and as decoded by linear probes on Wikipedia2Vec's word
  vectors averaged over the name, on its static entity vectors, and on Llama-2-7B (layer 24), as in
  \citet{gurnee2024language}'s Figure~1 (which shows Llama-2-70B); color marks
  the true continent. Panel titles give $R^2$ for latitude / longitude over all places; four places in Antarctica are not drawn. The entity vectors recover a blurred
  version of the layout that Llama-2 recovers more precisely (median error 2{,}450\,km against
  760\,km; Table~\ref{tab:entity}).}
  \label{fig:entity_map}
\end{figure}

\begin{table}[t]
  \centering
  \caption{Transformers and static embeddings on the entities released by
  \citet{gurnee2024language}, using their train--test split and scoring all
  representations on the held-out items covered by GloVe ($n = 6{,}690$ places, $7{,}316$
  figures); static values therefore differ slightly from Table~\ref{tab:gt}. Within-group
  $r$ correlates predictions with targets after removing group means in the test set.
  ``Input mean'' averages the model's input embeddings over all tokens of the name.
  ``Selected'' is the layer chosen on validation data within
  the training partition (Section~\ref{sec:llm_methods}).}
  \label{tab:gt_pythia}
  \footnotesize
  \begin{tabular}{lcccccc}
    \toprule
    & \multicolumn{3}{c}{Places (latitude / longitude)} & \multicolumn{3}{c}{Figures (year of death)} \\
    \cmidrule(lr){2-4} \cmidrule(lr){5-7}
    & & \multicolumn{2}{c}{Within-group $r$} & & \multicolumn{2}{c}{Within-group $r$} \\
    \cmidrule(lr){3-4} \cmidrule(lr){6-7}
    Representation & $R^2$ & Continent & Country & $R^2$ & Century & Decade \\
    \midrule
    GloVe                  & .47 / .53 & .33 / .14 & .06 / .05 & .50 & .05 & .00 \\
    Word2Vec               & .54 / .57 & .39 / .18 & .08 / .06 & .42 & .05 & .01 \\
    Pythia-2.8B, layer 0   & .50 / .57 & .36 / .15 & .11 / .07 & .34 & .06 & .00 \\
    Pythia-2.8B, selected  & .76 / .79 & .64 / .33 & .20 / .12 & .70 & .10 & .03 \\
    \quad 300 PCs          & .74 / .78 & .61 / .31 & .18 / .10 & .67 & .10 & .02 \\
    Pythia-2.8B, input mean & .52 / .60 & .39 / .16 & .10 / .06 & .52 & .08 & .00 \\
    Llama-2-7B, layer 0    & .49 / .56 & .36 / .14 & .10 / .05 & .31 & .05 & .01 \\
    Llama-2-7B, input mean & .53 / .60 & .40 / .17 & .10 / .08 & .52 & .05 & .00 \\
    Llama-2-7B, selected   & .88 / .88 & .80 / .50 & .35 / .23 & .78 & .12 & .02 \\
    \quad 300 PCs          & .86 / .87 & .77 / .45 & .29 / .18 & .76 & .11 & .02 \\
    \midrule
    Country / century mean (oracle) & .96 / .97 & .92 / .82 & --- & 1.00 & --- & --- \\
    \bottomrule
  \end{tabular}
\end{table}

\paragraph{Disambiguated static vectors.}
Averaged word vectors cannot resolve which entity a name refers to, and part of the
transformers' advantage could lie there. Static entity vectors, which carry only the associations
of an already disambiguated entity, close most of the gap (Table~\ref{tab:entity}, Figure~\ref{fig:entity_map}). On the items
they cover they reach $R^2 = 0.79$ on both coordinates, against $0.61$ and $0.66$ for
Wikipedia2Vec's own word vectors averaged over the name; this matches Pythia-2.8B (difference
$0.00$, 95\% interval $-0.09$ to $0.05$, resampling countries) and falls $0.11$ short of
Llama-2 ($0.07$ to $0.24$). On year of death they reach $0.80$, above Pythia ($+0.07$; $0.02$ to
$0.15$, resampling centuries) and indistinguishable from Llama-2 ($0.82$; difference $0.01$,
$-0.04$ to $0.05$), including in ordering figures within centuries. Llama-2's remaining advantage
on places is mostly precision: half its predictions fall within 760\,km of the true location,
against 2{,}450\,km for the entity vectors, and it orders places within countries better
($r = 0.36$ against $0.25$ for latitude). The entity vectors are given each entity's identity and are trained on Wikipedia, the source of the datasets (Section~\ref{sec:discussion}).

\begin{table}[t]
  \centering
  \caption{Word-level and entity-level static vectors against the transformers on the released
  entities that have a Wikipedia2Vec entity vector (authors' split; $n_{\text{test}} = 5{,}984$
  places and $4{,}707$ figures). Within-group $r$ is within country for places and within century
  for figures; the median error is in kilometers or years. Transformers at the selected layer.}
  \label{tab:entity}
  \footnotesize
  \setlength{\tabcolsep}{4pt}
  \begin{tabular}{lcccccc}
    \toprule
    & \multicolumn{3}{c}{Places} & \multicolumn{3}{c}{Figures} \\
    \cmidrule(lr){2-4} \cmidrule(lr){5-7}
    Representation & $R^2$ (lat / lon) & Within $r$ & Median error & $R^2$ & Within $r$ & Median error \\
    \midrule
    GloVe, word average          & .48 / .55 & .07 / .05 & 3{,}480 & .54 & .04 & 175 \\
    fastText, word average       & .59 / .62 & .12 / .08 & 3{,}038 & .62 & .07 & 160 \\
    Wikipedia2Vec, word average  & .61 / .66 & .13 / .09 & 2{,}879 & .65 & .07 & 150 \\
    Wikipedia2Vec, entity        & .79 / .79 & .25 / .11 & 2{,}448 & .80 & .14 & 109 \\
    \midrule
    Pythia-2.8B                  & .77 / .81 & .21 / .13 & 1{,}192 & .74 & .10 & 124 \\
    Llama-2-7B                   & .89 / .90 & .36 / .24 & 758     & .82 & .12 & 84 \\
    \bottomrule
  \end{tabular}
\end{table}

\subsection{What the Recovered Structure Consists Of}
\label{sec:mechanism-summary}

We fit a linear map from the static embeddings to the transformer activations on training
rows and probe the predicted part and the residual separately (Appendix~\ref{sec:residual_methods}).
On the released entities the static-predictable part takes out a third or less of the
activation variance but most of the probe's performance: removing it lowers $R^2$ on latitude
and longitude from $0.76$ and $0.79$ to $0.20$--$0.29$ for Pythia-2.8B and from $0.88$ to
$0.27$--$0.39$ for Llama-2-7B, and on year of death from $0.70$ to $0.21$--$0.25$ and from
$0.78$ to $0.29$--$0.33$, depending on the static model. Removing the same share of variance
along random directions, or static vectors assigned to the wrong entities, leaves performance
almost unchanged.

Because both kinds of representation carry country and century identity, this cannot show
whether what they share goes beyond group membership. We therefore also removed from the
activations everything linearly predictable from a one-hot code of each item's country or
century. This takes out a fifth to a quarter of the activation variance for places and under
4\% for figures, but nearly all of the probe's performance ($R^2 \leq 0.01$ on every target
in both models). A probe trained on what remains still orders items within their group
(Llama-2: $r = 0.55$ and $0.53$ within countries for latitude and longitude, $0.13$ within
centuries; Pythia: $0.47$, $0.49$ and $0.09$). Removing in addition the static structure that
goes beyond group membership lowers these to $0.46$, $0.39$ and $0.03$ for Llama-2 (Pythia:
$0.37$, $0.34$ and $0.00$), against $0.54$, $0.52$ and $0.13$ (Pythia: $0.46$, $0.49$ and
$0.09$) after removing a random subspace of the same variance; for every target in both models
the 95\% interval for this difference excludes zero, and appending the static vectors to the
activations raises $R^2$ by at most $0.01$ (Table~\ref{tab:groupremoval}, Appendix~\ref{sec:group_methods}).
What the two kinds of representation share is thus mostly coarse group membership, plus part of the within-country ordering of places and the transformers' weak ordering of figures within centuries.

Finally, a random split tests generalization only to new items from familiar groups. When
whole groups are held out, no representation we tested generalizes well: with each continent
held out in turn, every representation predicts the held-out coordinates worse than their mean
($R^2$ from $-0.94$ to $-0.22$ for the word vectors and transformers and $-0.03$ and $-0.68$ for the entity vectors, against $0.48$--$0.89$ under random splits), and with each block of death
years held out $R^2$ is $-0.27$ for the entity vectors and $-0.16$ and $-0.11$ for Word2Vec and GloVe but $0.11$ and $0.19$ for
Pythia and Llama-2 (Table~\ref{tab:grouped_holdout}).

\subsection{Which Words Carry the Signal}
\label{sec:words}

\begin{figure}[t]
  \centering
  \includegraphics[width=0.92\textwidth]{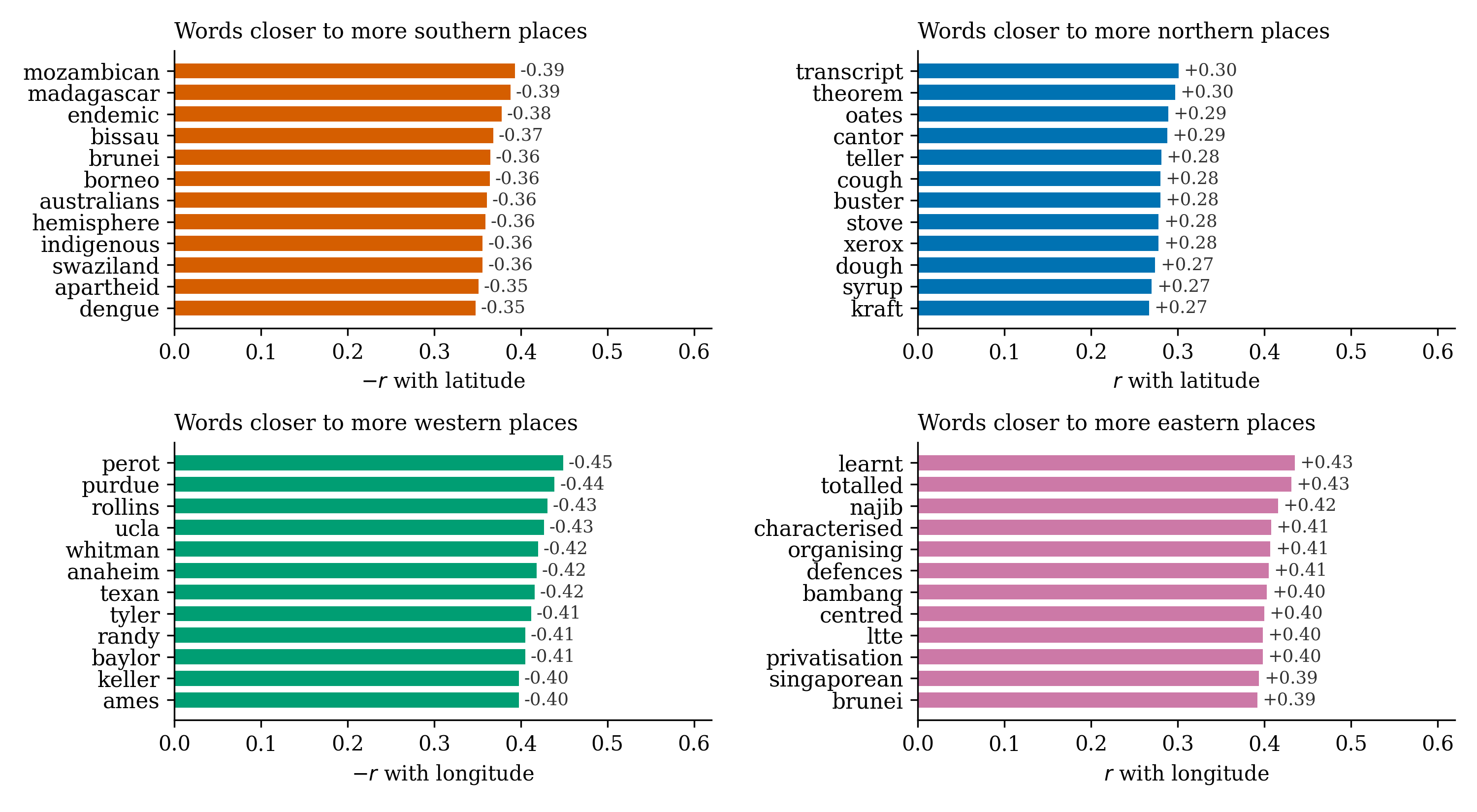}
  \caption{Words whose GloVe vectors are most similar to place names that lie farther south, north, west or east. For each of 15{,}923 words (the 20{,}000 most frequent GloVe words with at least four letters, excluding a list of country names, demonyms and geographic terms and any word that is part of at least three place names), we correlate the word's cosine similarity to the averaged name vector of each of the 33{,}500 released places that have a GloVe vector with the place's latitude or longitude; shown are the twelve strongest correlations at each end. The words are selected for extreme correlations from the whole vocabulary, so the figure is descriptive and implies no significance.}
  \label{fig:word_coords}
\end{figure}

To see what a probe on static vectors could be reading, we asked which words are most similar to the names of places that lie farther south or north, west or east. For each frequent word we correlated its cosine similarity to the averaged name vector of every released place with the place's latitude and longitude (Figure~\ref{fig:word_coords}).

The longitude words are the most legible. Eastern places are closer to British spellings (\emph{learnt}, \emph{organising}, \emph{centred}, \emph{privatisation}), a convention of British rather than American English, and western places to words tied to American universities and cities (\emph{ucla}, \emph{purdue}, \emph{baylor}, \emph{anaheim}). At the southern end of latitude are African, Australian and tropical terms (\emph{mozambican}, \emph{madagascar}, \emph{australians}, \emph{dengue}); the northern end mixes academic words (\emph{transcript}, \emph{theorem}) with the names of companies and politicians (\emph{xerox}, \emph{caterpillar}, \emph{hastert}), and we do not interpret it. The point is not that every word is interpretable, but that many individual words track a place's coordinates at correlations up to $0.45$ although none of them states a coordinate. This is consistent with the coarse sorting by country and region described above (Section~\ref{sec:mechanism-summary}), available from how words are used and with no context-dependent computation over the name. Projecting out the embedding directions spanned by country names costs the probe more than projecting out directions spanned by matched random words (Appendix~\ref{sec:semantic_ablation}).

\subsection{A ``Pain Axis''}
\label{sec:pain}

\citet{tagliabue2026pain} extract a ``pain direction'' from 25 open-weight models (2B to
72B parameters, five families). They report that it separates descriptions of painful
situations from matched controls, retains a component distinct from fear and negative valence
after their shared variance is removed, separates pain from injuries described as painless,
and, when added to the residual stream, produces first-person expressions of worthlessness
and failure. We reran their representational analyses with static word
embeddings (Section~\ref{sec:pain_methods}).

\begin{table}[t]
  \centering
  \caption{The pain-axis tests on static word embeddings, using the v2 stimuli of
  \citet{tagliabue2026pain}. Held-out AUC for pain against the S2 or S1
  controls and, for the S2 direction, the standalone Arousal and Random sets (mean $\pm$ s.d.\ over 20 fold assignments);
  $z$: mean projection of the standalone Numb and Sadness sets on the S2 direction,
  referenced to the S2 sentences (pain $\approx +0.7$, controls $\approx -0.7$). LLM values are the
  authors' held-out estimates; the keyword baseline is from \citet{painreview2026}.}
  \label{tab:pain}
  \small
  \begin{tabular}{lcccccc}
    \toprule
    & \multicolumn{4}{c}{Held-out AUC} & \multicolumn{2}{c}{$z$ on S2 direction} \\
    \cmidrule(lr){2-5} \cmidrule(lr){6-7}
    Representation & S2 & S1 & S2 vs.\ Arousal & S2 vs.\ Random & Numb & Sadness \\
    \midrule
    Keyword baseline & $0.727$ & $0.669$ & --- & --- & --- & --- \\
    GloVe    & $0.850 \pm 0.009$ & $0.773 \pm 0.011$ & $0.664$ & $0.815$ & $-0.07$ & $-0.09$ \\
    Word2Vec & $0.867 \pm 0.007$ & $0.802 \pm 0.012$ & $0.693$ & $0.857$ & $-0.38$ & $-0.48$ \\
    fastText & $0.878 \pm 0.006$ & $0.785 \pm 0.013$ & $0.712$ & $0.883$ & $-0.67$ & $-0.50$ \\
    \midrule
    25 LLMs  & $0.91$--$1.00$ & $0.85$--$0.94$ & all separate\textsuperscript{a} & all separate\textsuperscript{a} & $-0.4$--$+0.3$ & --- \\
    \bottomrule
    \multicolumn{7}{l}{\footnotesize \textsuperscript{a}Reported to separate pain from these sets in every model; values not given.}
  \end{tabular}

\end{table}

\paragraph{Separation.}
Averaged word vectors separate pain from controls on held-out sentence sets at AUC
$0.85$--$0.88$ on S2 and $0.77$--$0.80$ on S1 (Table~\ref{tab:pain}; see also
Figure~\ref{fig:pain} in the appendix). An independent review tested surface-lexical
baselines under the same
cross-validation, found at best $0.73$ and $0.67$, and concluded that the separation is not
lexical \citep{painreview2026}. That holds for surface features, but static embeddings
close about half the gap between keyword counts and the LLMs. The LLMs are still better on
held-out sets ($0.91$--$1.00$ on S2, $0.85$--$0.94$ on S1), even though their values come
from the best of many layers. The standalone sets show a clearer gap. The
static S2 direction separates held-out pain sentences from the Random set of everyday
statements (AUC $0.82$--$0.88$) but only weakly from the Arousal set of intense positive
experiences ($0.66$--$0.71$), and the S1 direction separates pain from neither
($0.44$--$0.60$), whereas the LLM directions separate pain from both in every model.

\paragraph{Specificity.}
The case that the direction encodes pain in particular, rather than distress in general,
originally rested on its low cosines with fear and negative emotion. In their second, more recent
version, the authors show that these values depend on how the directions are built: the pain
direction subtracts the pooled controls, which contain the fear and negative-emotion
sentences, so it is partly orthogonalized against them by construction. Built against a
common neutral baseline, the S2 direction's cosine with fear rises from $+0.12$ to $+0.58$
and with negative emotion from $+0.21$ to $+0.77$; built against the pooled controls, the
values are $+0.08$ and $+0.14$. We find that static embeddings reproduce this pattern under all three
constructions (Table~\ref{tab:pain_cos}, appendix). Both the large component that pain shares
with fear and negative emotion and the small residual that survives a common control
baseline are therefore present in averaged word vectors. The models differ in the residual
overlap with sadness, the largest the authors report ($+0.38$); in static embeddings it is
$+0.10$ to $+0.19$.

Injuries described as painless project well below the pain sentences ($z = -0.07$ to $-0.67$, against about $+0.7$), as in the models.

\paragraph{Summary.}
Averaged word vectors also separate pain from matched controls and show the same
construction-dependent overlap with fear and negative emotion. These tests reflect the
distributional structure of stimuli written to differ in exactly these ways and
do not by themselves show that the models represent pain. The models go further on held-out
separation, separation from arousal, the overlap with sadness and the vocabulary the direction
promotes (\emph{hurt}, \emph{shame}, \emph{worthless}); the
behavioral results lie outside what a representational baseline can test.

\subsection{Emotion Concepts: A Conceptual Replication}
\label{sec:emotion}

\citet{sofroniew2026emotion} report that emotion directions in Claude Sonnet 4.5 generalize
across contexts and causally influence the model's outputs (which they call \emph{functional
emotions}, without implying subjective experience). We test the representational claim that the
directions encode emotion concepts rather than emotion words, for which the evidence offered is
a prompt forbidding the story writer to use the emotion word or its synonyms
(Section~\ref{sec:emotion_methods}); since the original stories are not released, this tests
the protocol rather than Claude on the same stories. The prohibition removes the word but not the
vocabulary that goes with it. The authors' example for \emph{desperate} (their Appendix 6.6), a man facing arrest, out of a job and left by his wife, contains none of that word but many of the words found near it in text.

\paragraph{Results.}
On held-out topics, averaged word vectors separate each emotion from the other eleven at a
mean AUC of $0.84$--$0.88$, and assign stories to the right emotion 41--51\% of the time
(balanced accuracy, the mean recall over emotions, 44--54\%), against 8.3\% for a uniform
guess and 11.2\% for always choosing the most frequent held-out emotion (Table~\ref{tab:emotion}). The word directions, built from the
emotion labels alone, reach a mean AUC of $0.75$--$0.80$ and 12-way accuracy of 28--33\%
(balanced 29--35\%), three to four times chance. The story directions are reliably better than the word
directions (by $0.04$--$0.13$ in AUC and $0.06$--$0.28$ in accuracy, 95\% intervals over
held-out topics), but a substantial share of what they recover is already present in the
static vectors for the emotion words.

Because these stories were written to satisfy the prohibition, we also tested it on text
that was not. On GoEmotions, mean held-out AUC across the 27 emotions is $0.79$--$0.82$.
After removing every comment that contains an emotion word or a synonym, it is $0.78$--$0.81$,
a drop of about one point in each embedding. Most of the signal does not come from the
emotion words themselves.

\begin{table}[t]
  \centering
  \caption{Emotion directions on averaged static word vectors. Stories come from an
  independent replication corpus \citep{ogunlana2026qwenemotion} generated under the prompt
  of \citet{sofroniew2026emotion}; topics are held out, and brackets give 95\% intervals from
  resampling held-out topics. The word direction uses the emotion label's static vector and no
  stories. GoEmotions \citep{demszky2020goemotions} has 27 emotions in human-written text; the
  last column excludes every comment containing an emotion word or synonym.}
  \label{tab:emotion}
  \small
  \begin{tabular}{lccccc}
    \toprule
    & \multicolumn{2}{c}{Replication stories (12 emotions)} & & \multicolumn{2}{c}{GoEmotions (27 emotions)} \\
    \cmidrule(lr){2-3} \cmidrule(lr){5-6}
    & AUC & 12-way acc. & & all text & emotion words removed \\
    \midrule
    GloVe, story direction    & $0.838$ {\scriptsize [.82, .86]} & $0.415$ {\scriptsize [.36, .48]} & & $0.791$ & $0.783$ \\
    GloVe, word direction     & $0.752$ {\scriptsize [.72, .78]} & $0.277$ {\scriptsize [.24, .32]} & & --- & --- \\
    Word2Vec, story direction & $0.873$ {\scriptsize [.85, .89]} & $0.484$ {\scriptsize [.41, .55]} & & $0.804$ & $0.794$ \\
    Word2Vec, word direction  & $0.779$ {\scriptsize [.75, .81]} & $0.330$ {\scriptsize [.27, .39]} & & --- & --- \\
    fastText, story direction & $0.876$ {\scriptsize [.85, .90]} & $0.511$ {\scriptsize [.45, .57]} & & $0.818$ & $0.811$ \\
    fastText, word direction  & $0.802$ {\scriptsize [.77, .83]} & $0.314$ {\scriptsize [.24, .38]} & & --- & --- \\
    \midrule
    Chance                    & $0.500$ & $0.083$ & & $0.500$ & $0.500$ \\
       \bottomrule
  \end{tabular}
\end{table}

\paragraph{Summary.}
Averaged static word vectors reproduce the evidence that emotion concepts can be decoded, that
decoding generalizes to new topics, and that it survives a ban on naming the emotion, and the
vectors for the emotion words alone reproduce most of it. These tests reflect the vocabulary of
stimuli written to differ in exactly this respect and do not on their own show
that a model represents emotion concepts in a stronger sense than a lexicon does. The authors'
graded scenarios, in which one quantity varies while the wording stays
nearly constant, cannot be tested with averaged word vectors, and the steering results lie
outside what a representational baseline can address; these are where evidence about the model
itself is most likely to be found.

\section{Discussion}
\label{sec:discussion}

The case for emergent spatial and temporal world models has rested mainly on three observations: that probes recover real-world variables accurately, that accuracy rises across layers, and that it extends to entities held out from probe training. Static embeddings recover a large share of what has been decoded: averaged word vectors explain about half the variance in place coordinates and year of death on the released datasets, and static vectors for disambiguated entities reach $R^2 = 0.79$ and $0.80$, matching Pythia-2.8B on coordinates and Llama-2-7B on year of death. Held-out accuracy holds for static embeddings too, and when whole continents or periods are held out no representation we tested generalizes well (Section~\ref{sec:mechanism-summary}). High accuracy alone therefore cannot distinguish a representation of the world from the distributional associations in the training text.

Language models do better than this baseline. On the items that have an entity vector, Llama-2-7B reaches $R^2 = 0.89$ and $0.90$ on coordinates, against $0.79$ for entity vectors, and its median error is 758~km against 2{,}448~km (Table~\ref{tab:entity}). However, much of that advantage lies in sorting items into the right group, the right country or the right century. A predictor that knows only each item's group already captures nearly all of the variance ($R^2 = 0.96$--$1.00$), while within groups the language models' gain is modest (Table~\ref{tab:gt_pythia}). Group membership is also the kind of information word associations carry: places in the same country share a language, a currency, institutions and neighboring places, and people from the same century share events and vocabulary. Ordering items \emph{within} a group, or placing a country relative to a distant one, is what a map or a timeline should support, and that is where the advantage over the baseline is smallest. Part of the rise in accuracy across layers reflects the model assembling multi-token names, but averaging a name's input embeddings does not close the gap, so most of it comes from processing in context; the added accuracy is again mostly placement in the right country or century (Table~\ref{tab:gt_pythia}).

We are not claiming, on the basis of these results, that language models necessarily lack internal models of the world. The language models' advantage over the baseline could reflect a more sophisticated encoding of distributional associations alone, or something of a different kind, and our comparisons cannot fully separate the two. The entity baseline should not be read as a capacity-matched competitor to the transformer: it is given each entity's identity, which a transformer must infer from the name, and it is trained on Wikipedia, the source of the datasets. It is an existence proof: once the correct referent is supplied, fixed distributional associations attached to that entity suffice to reproduce most of the probe performance. Our claim is simply that the evidence offered so far, decoding accuracy and generalization to held-out entities, does not by itself distinguish a representation of the world from distributional associations.

The baseline suggests a simple criterion for future work. Any decoding result should be reported against static embeddings on the same items, and the advantage should be characterized in two ways: its size, and its kind, that is, whether it is a finer sorting of the same type that associations provide or a structure of a different type. Distinguishing these requires tests that set the property in question against the words used, for example items whose described location or date differs from what their names and neighbors would suggest, or groups held out of training entirely (Section~\ref{sec:mechanism-summary} and Appendix~\ref{sec:grouped_holdout}); causal and behavioral effects, which no representational baseline can address, are a further source of evidence. Without such tests, a probe result is consistent with both readings. The baseline is cheap, needs no access to the model, and can be applied to any study in this literature. It is also a reminder of how much the company a word keeps \citep{firth1957synopsis} records about where things are, when they happened, and how situations feel to the people who describe them.

\paragraph{Limitations.}
The largest transformer we tested is Llama-2-7B, and the
strongest claims concern much larger models, which may carry more than coarse sorting or generalize
better to held-out continents or periods; the baseline should be run against them. Our comparisons are representational, so they do not
address causal and behavioral evidence, such as steering or the Othello interventions \citep{li2023emergent}, or whether a model uses what a
decoder can read. The baseline itself is deliberately simple: averaging word vectors
ignores word order, so it likely understates what word relations alone can provide, and our
separation of what the transformers share with it is linear. Items with an entity vector are easier for every representation, language models included (Appendix~\ref{sec:coverage}), so absolute accuracy on the common subset overstates accuracy on the full release. The emotion case rests on a
replication corpus rather than the original stories.

\paragraph{Broader impact.}
Decoding and related representational results are increasingly discussed in connection with model welfare and sentience \citep{keeling2024tradeoffs}. Our
results caution against treating decodability as evidence of internal states, but do not
show that models lack such states; we hope the baseline makes such discussions better calibrated rather than dismissive.

\paragraph{Reproducibility.}
Code, derived data and the result files behind every reported value are available
\ifanon in the supplementary material\else at \url{https://github.com/elanbarenholtz/static-embeddings-space-time}\fi,
including \texttt{static\_baseline.py}, which runs the baseline
on any labeled stimulus set. The entity analyses use the 300-dimensional English Wikipedia2Vec vectors (April 2018 release), which the supplementary scripts filter to the items used.\ifanon\else{} An earlier version of this work also analyzed a set of 272 cities; this version removes that analysis, and all results here fit an intercept.\fi

\bibliography{references}

\begin{thebibliography}{45}
\providecommand{\natexlab}[1]{#1}
\providecommand{\url}[1]{\texttt{#1}}
\expandafter\ifx\csname urlstyle\endcsname\relax
  \providecommand{\doi}[1]{doi: #1}\else
  \providecommand{\doi}{doi: \begingroup \urlstyle{rm}\Url}\fi

\bibitem[Alain and Bengio(2016)]{alain2016understanding}
Guillaume Alain and Yoshua Bengio.
\newblock Understanding intermediate layers using linear classifier probes.
\newblock \emph{arXiv preprint arXiv:1610.01644}, 2016.

\bibitem[Azaria and Mitchell(2023)]{azaria2023internal}
Amos Azaria and Tom Mitchell.
\newblock The internal state of an {LLM} knows when it's lying.
\newblock In \emph{Findings of the Association for Computational Linguistics:
  EMNLP 2023}, 2023.

\bibitem[Belinkov(2022)]{belinkov2022probing}
Yonatan Belinkov.
\newblock Probing classifiers: Promises, shortcomings, and advances.
\newblock \emph{Computational Linguistics}, 48\penalty0 (1):\penalty0 207--219,
  2022.

\bibitem[Biderman et~al.(2023)Biderman, Schoelkopf, Anthony, Bradley, O'Brien,
  Hallahan, Khan, Purohit, Prashanth, Raff, Skowron, Sutawika, and van~der
  Wal]{biderman2023pythia}
Stella Biderman, Hailey Schoelkopf, Quentin Anthony, Herbie Bradley, Kyle
  O'Brien, Eric Hallahan, Mohammad~Aflah Khan, Shivanshu Purohit, USVSN~Sai
  Prashanth, Edward Raff, Aviya Skowron, Lintang Sutawika, and Oskar van~der
  Wal.
\newblock Pythia: A suite for analyzing large language models across training
  and scaling.
\newblock In \emph{Proceedings of the 40th International Conference on Machine
  Learning}, pages 2397--2430, 2023.

\bibitem[Bojanowski et~al.(2017)Bojanowski, Grave, Joulin, and
  Mikolov]{bojanowski2017enriching}
Piotr Bojanowski, Edouard Grave, Armand Joulin, and Tomas Mikolov.
\newblock Enriching word vectors with subword information.
\newblock \emph{Transactions of the Association for Computational Linguistics},
  5:\penalty0 135--146, 2017.

\bibitem[Bolukbasi et~al.(2016)Bolukbasi, Chang, Zou, Saligrama, and
  Kalai]{bolukbasi2016man}
Tolga Bolukbasi, Kai-Wei Chang, James~Y. Zou, Venkatesh Saligrama, and Adam~T.
  Kalai.
\newblock Man is to computer programmer as woman is to homemaker? {D}ebiasing
  word embeddings.
\newblock \emph{Advances in Neural Information Processing Systems}, 29, 2016.

\bibitem[Burns et~al.(2023)Burns, Ye, Klein, and
  Steinhardt]{burns2023discovering}
Collin Burns, Haotian Ye, Dan Klein, and Jacob Steinhardt.
\newblock Discovering latent knowledge in language models without supervision.
\newblock In \emph{International Conference on Learning Representations}, 2023.

\bibitem[Conneau et~al.(2018)Conneau, Kruszewski, Lample, Barrault, and
  Baroni]{conneau2018you}
Alexis Conneau, German Kruszewski, Guillaume Lample, Lo{\"\i}c Barrault, and
  Marco Baroni.
\newblock What you can cram into a single \$\&!\#* vector: Probing sentence
  embeddings for linguistic properties.
\newblock In \emph{Proceedings of the 56th Annual Meeting of the Association
  for Computational Linguistics}, pages 2126--2136, 2018.

\bibitem[Demszky et~al.(2020)Demszky, Movshovitz-Attias, Ko, Cowen, Nemade, and
  Ravi]{demszky2020goemotions}
Dorottya Demszky, Dana Movshovitz-Attias, Jeongwoo Ko, Alan Cowen, Gaurav
  Nemade, and Sujith Ravi.
\newblock {GoEmotions}: A dataset of fine-grained emotions.
\newblock In \emph{Proceedings of the 58th Annual Meeting of the Association
  for Computational Linguistics}, pages 4040--4054, 2020.

\bibitem[Elazar et~al.(2021)Elazar, Ravfogel, Jacovi, and
  Goldberg]{elazar2021amnesic}
Yanai Elazar, Shauli Ravfogel, Alon Jacovi, and Yoav Goldberg.
\newblock Amnesic probing: Behavioral explanation with amnesic counterfactuals.
\newblock \emph{Transactions of the Association for Computational Linguistics},
  9:\penalty0 160--175, 2021.

\bibitem[Farquhar et~al.(2023)Farquhar, Varma, Kenton, Gasteiger, Mikulik, and
  Shah]{farquhar2023challenges}
Sebastian Farquhar, Vikrant Varma, Zachary Kenton, Johannes Gasteiger, Vladimir
  Mikulik, and Rohin Shah.
\newblock Challenges with unsupervised {LLM} knowledge discovery.
\newblock \emph{arXiv preprint arXiv:2312.10029}, 2023.

\bibitem[Firth(1957)]{firth1957synopsis}
John~Rupert Firth.
\newblock A synopsis of linguistic theory, 1930--1955.
\newblock In \emph{Studies in Linguistic Analysis}, pages 1--32. Blackwell,
  Oxford, 1957.

\bibitem[Grand et~al.(2022)Grand, Blank, Pereira, and
  Fedorenko]{grand2022semantic}
Gabriel Grand, Idan~Asher Blank, Francisco Pereira, and Evelina Fedorenko.
\newblock Semantic projection recovers rich human knowledge of multiple object
  features from word embeddings.
\newblock \emph{Nature Human Behaviour}, 6\penalty0 (7):\penalty0 975--987,
  2022.

\bibitem[Gupta et~al.(2015)Gupta, Boleda, Baroni, and
  Pad{\'o}]{gupta2015distributional}
Abhijeet Gupta, Gemma Boleda, Marco Baroni, and Sebastian Pad{\'o}.
\newblock Distributional vectors encode referential attributes.
\newblock In \emph{Proceedings of the 2015 Conference on Empirical Methods in
  Natural Language Processing}, pages 12--21, 2015.

\bibitem[Gurnee and Tegmark(2024)]{gurnee2024language}
Wes Gurnee and Max Tegmark.
\newblock Language models represent space and time.
\newblock In \emph{International Conference on Learning Representations}, 2024.

\bibitem[Heinzerling and Inui(2024)]{heinzerling2024monotonic}
Benjamin Heinzerling and Kentaro Inui.
\newblock Monotonic representation of numeric properties in language models.
\newblock In \emph{Proceedings of the 62nd Annual Meeting of the Association
  for Computational Linguistics (Volume 2: Short Papers)}, 2024.

\bibitem[Hewitt and Liang(2019)]{hewitt2019designing}
John Hewitt and Percy Liang.
\newblock Designing and interpreting probes with control tasks.
\newblock In \emph{Proceedings of the 2019 Conference on Empirical Methods in
  Natural Language Processing}, pages 2733--2743, 2019.

\bibitem[Hewitt and Manning(2019)]{hewitt2019structural}
John Hewitt and Christopher~D. Manning.
\newblock A structural probe for finding syntax in word representations.
\newblock In \emph{Proceedings of the 2019 Conference of the North {A}merican
  Chapter of the Association for Computational Linguistics: Human Language
  Technologies}, pages 4129--4138, 2019.

\bibitem[Hoerl and Kennard(1970)]{hoerl1970ridge}
Arthur~E. Hoerl and Robert~W. Kennard.
\newblock Ridge regression: Biased estimation for nonorthogonal problems.
\newblock \emph{Technometrics}, 12\penalty0 (1):\penalty0 55--67, 1970.

\bibitem[Hollis et~al.(2017)Hollis, Westbury, and
  Lefsrud]{hollis2017extrapolating}
Geoff Hollis, Chris Westbury, and Lianne Lefsrud.
\newblock Extrapolating human judgments from skip-gram vector representations
  of word meaning.
\newblock \emph{Quarterly Journal of Experimental Psychology}, 70\penalty0
  (8):\penalty0 1603--1619, 2017.

\bibitem[Karvonen(2024)]{karvonen2024emergent}
Adam Karvonen.
\newblock Emergent world models and latent variable estimation in chess-playing
  language models.
\newblock In \emph{Conference on Language Modeling (COLM)}, 2024.

\bibitem[Keeling et~al.(2024)]{keeling2024tradeoffs}
Geoff Keeling et~al.
\newblock Can {LLMs} make trade-offs involving stipulated pain and pleasure
  states?
\newblock \emph{arXiv preprint arXiv:2411.02432}, 2024.

\bibitem[Konkol et~al.(2017)Konkol, Brychc{\'\i}n, Nykl, and
  Hercig]{konkol2017geographical}
Michal Konkol, Tom{\'a}{\v{s}} Brychc{\'\i}n, Michal Nykl, and Tom{\'a}{\v{s}}
  Hercig.
\newblock Geographical evaluation of word embeddings.
\newblock In \emph{Proceedings of the Eighth International Joint Conference on
  Natural Language Processing (Volume 1: Long Papers)}, pages 224--232, 2017.

\bibitem[Kozlowski et~al.(2019)Kozlowski, Taddy, and
  Evans]{kozlowski2019geometry}
Austin~C. Kozlowski, Matt Taddy, and James~A. Evans.
\newblock The geometry of culture: Analyzing the meanings of class through word
  embeddings.
\newblock \emph{American Sociological Review}, 84\penalty0 (5):\penalty0
  905--949, 2019.

\bibitem[Levinstein and Herrmann(2024)]{levinstein2024still}
Benjamin~A. Levinstein and Daniel~A. Herrmann.
\newblock Still no lie detector for language models: Probing empirical and
  conceptual roadblocks.
\newblock \emph{Philosophical Studies}, 2024.
\newblock \doi{10.1007/s11098-023-02094-3}.

\bibitem[Li et~al.(2023)Li, Hopkins, Bau, Vi{\'e}gas, Pfister, and
  Wattenberg]{li2023emergent}
Kenneth Li, Aspen~K. Hopkins, David Bau, Fernanda Vi{\'e}gas, Hanspeter
  Pfister, and Martin Wattenberg.
\newblock Emergent world representations: Exploring a sequence model trained on
  a synthetic task.
\newblock In \emph{International Conference on Learning Representations
  (ICLR)}, 2023.

\bibitem[Li{\'e}tard et~al.(2021)Li{\'e}tard, Abdou, and
  S{\o}gaard]{lietard2021rome}
Bastien Li{\'e}tard, Mostafa Abdou, and Anders S{\o}gaard.
\newblock Do language models know the way to {R}ome?
\newblock In \emph{Proceedings of the Fourth BlackboxNLP Workshop on Analyzing
  and Interpreting Neural Networks for NLP}, pages 510--517, 2021.

\bibitem[Louwerse and Zwaan(2009)]{louwerse2009language}
Max~M. Louwerse and Rolf~A. Zwaan.
\newblock Language encodes geographical information.
\newblock \emph{Cognitive Science}, 33\penalty0 (1):\penalty0 51--73, 2009.

\bibitem[Marks and Tegmark(2023)]{marks2024geometry}
Samuel Marks and Max Tegmark.
\newblock The geometry of truth: Emergent linear structure in large language
  model representations of true/false datasets.
\newblock \emph{arXiv preprint arXiv:2310.06824}, 2023.

\bibitem[Mikolov et~al.(2013{\natexlab{a}})Mikolov, Chen, Corrado, and
  Dean]{mikolov2013efficient}
Tomas Mikolov, Kai Chen, Greg Corrado, and Jeffrey Dean.
\newblock Efficient estimation of word representations in vector space.
\newblock \emph{arXiv preprint arXiv:1301.3781}, 2013{\natexlab{a}}.

\bibitem[Mikolov et~al.(2013{\natexlab{b}})Mikolov, Yih, and
  Zweig]{mikolov2013linguistic}
Tomas Mikolov, Wen-tau Yih, and Geoffrey Zweig.
\newblock Linguistic regularities in continuous space word representations.
\newblock In \emph{Proceedings of the 2013 Conference of the North American
  Chapter of the Association for Computational Linguistics: Human Language
  Technologies}, pages 746--751, 2013{\natexlab{b}}.

\bibitem[Nanda et~al.(2023)Nanda, Lee, and Wattenberg]{nanda2023emergent}
Neel Nanda, Andrew Lee, and Martin Wattenberg.
\newblock Emergent linear representations in world models of self-supervised
  sequence models.
\newblock In \emph{Proceedings of the 6th BlackboxNLP Workshop}, 2023.

\bibitem[Ogunlana(2026)]{ogunlana2026qwenemotion}
Foluso Ogunlana.
\newblock Qwen emotion stories: A replication corpus for ``{Emotion} concepts
  and their function in a large language model''.
\newblock Hugging Face dataset,
  \url{https://huggingface.co/datasets/foogunlana/qwen-emotion-stories}, 2026.
\newblock Revision \texttt{be8f6def898b}, accessed September 22, 2026.

\bibitem[{pain-axis-review contributors}(2026)]{painreview2026}
{pain-axis-review contributors}.
\newblock pain-axis-review: A reproducible review of {arXiv:2609.16247}.
\newblock \url{https://github.com/wolframs/pain-axis-review}, 2026.

\bibitem[Pennington et~al.(2014)Pennington, Socher, and
  Manning]{pennington2014glove}
Jeffrey Pennington, Richard Socher, and Christopher~D. Manning.
\newblock {GloVe}: Global vectors for word representation.
\newblock In \emph{Proceedings of the 2014 Conference on Empirical Methods in
  Natural Language Processing (EMNLP)}, pages 1532--1543, 2014.

\bibitem[Ravichander et~al.(2021)Ravichander, Belinkov, and
  Hovy]{ravichander2021probing}
Abhilasha Ravichander, Yonatan Belinkov, and Eduard Hovy.
\newblock Probing the probing paradigm: Does probing accuracy entail task
  relevance?
\newblock In \emph{Proceedings of the 16th Conference of the European Chapter
  of the Association for Computational Linguistics}, pages 3363--3377, 2021.

\bibitem[Sofroniew et~al.(2026)]{sofroniew2026emotion}
Nicholas Sofroniew et~al.
\newblock Emotion concepts and their function in a large language model.
\newblock \emph{arXiv preprint arXiv:2604.07729}, 2026.

\bibitem[Sun et~al.(2026)]{sun2026valence}
Lihao Sun et~al.
\newblock Valence--arousal subspace in {LLMs}: Circular emotion geometry and
  multi-behavioral control.
\newblock \emph{arXiv preprint arXiv:2604.03147}, 2026.

\bibitem[Tagliabue et~al.(2026)Tagliabue, Dung, and Berg]{tagliabue2026pain}
Valen Tagliabue, Leonard Dung, and Cameron Berg.
\newblock The pain axis: {LLMs} represent self-directed harm and act on it.
\newblock \emph{arXiv preprint arXiv:2609.16247}, 2026.
\newblock Version 2, 25 September 2026.

\bibitem[Tigges et~al.(2023)Tigges, Hollinsworth, Geiger, and
  Nanda]{tigges2023linear}
Curt Tigges, Oskar~John Hollinsworth, Atticus Geiger, and Neel Nanda.
\newblock Linear representations of sentiment in large language models.
\newblock \emph{arXiv preprint arXiv:2310.15154}, 2023.

\bibitem[Touvron et~al.(2023)Touvron, Martin, Stone, Albert, Almahairi, Babaei,
  Bashlykov, Batra, Bhargava, Bhosale, et~al.]{touvron2023llama}
Hugo Touvron, Louis Martin, Kevin Stone, Peter Albert, Amjad Almahairi, Yasmine
  Babaei, Nikolay Bashlykov, Soumya Batra, Prajjwal Bhargava, Shruti Bhosale,
  et~al.
\newblock Llama 2: Open foundation and fine-tuned chat models.
\newblock \emph{arXiv preprint arXiv:2307.09288}, 2023.

\bibitem[Vrande{\v{c}}i{\'c} and Kr{\"o}tzsch(2014)]{vrandecic2014wikidata}
Denny Vrande{\v{c}}i{\'c} and Markus Kr{\"o}tzsch.
\newblock {Wikidata}: A free collaborative knowledgebase.
\newblock \emph{Communications of the ACM}, 57\penalty0 (10):\penalty0 78--85,
  2014.

\bibitem[Yamada et~al.(2020)Yamada, Asai, Sakuma, Shindo, Takeda, Takefuji, and
  Matsumoto]{yamada2020wikipedia2vec}
Ikuya Yamada, Akari Asai, Jin Sakuma, Hiroyuki Shindo, Hideaki Takeda,
  Yoshiyasu Takefuji, and Yuji Matsumoto.
\newblock {Wikipedia2Vec}: An efficient toolkit for learning and visualizing
  the embeddings of words and entities from {Wikipedia}.
\newblock In \emph{Proceedings of the 2020 Conference on Empirical Methods in
  Natural Language Processing: System Demonstrations}, pages 23--30, 2020.

\bibitem[Zhu et~al.(2024)Zhu, Zhang, and Wang]{zhu2024language}
Wentao Zhu, Zhining Zhang, and Yizhou Wang.
\newblock Language models represent beliefs of self and others.
\newblock In \emph{Proceedings of the 41st International Conference on Machine
  Learning}, volume 235 of \emph{PMLR}, 2024.

\bibitem[Zou et~al.(2023)Zou, Phan, et~al.]{zou2023representation}
Andy Zou, Long Phan, et~al.
\newblock Representation engineering: A top-down approach to {AI} transparency.
\newblock \emph{arXiv preprint arXiv:2310.01405}, 2023.

\end{thebibliography}
\bibliographystyle{plainnat}

\appendix

\section{Supporting Analyses}
\label{sec:appendix}

The analyses in this appendix characterize \emph{why} the type-level baseline
recovers what it does, and report controls on the baseline itself. They support the
results in Section~\ref{sec:results} but are not required for the argument.

\subsection{Additional Methods}
\label{sec:extra_methods}

\paragraph{Activation extraction.} Activations were extracted in 16-bit precision and checked
against a separately computed CPU reference on 60 random names per model and dataset (cosine
similarity $> 0.999$).

\paragraph{Layer selection.} Single-layer results use a layer chosen without the test
data. On the released entities, the authors' training partition is split once into an
inner training set (80\%) and a validation set (20\%); each saved layer is probed on the
inner training set and scored on the validation set, and the layer with the best validation
$R^2$ (averaged over targets) is refitted on the full training partition and scored once on
the test partition. This selects layer~24 for Pythia-2.8B on places and layer~28 on
figures, and layer~24 for Llama-2-7B on both. Analyses carried out at a single layer (residualization, matched dimensionality) use the layer selected on the training partition, so no test item influences it. For the grouped holdouts, where the authors' training partition
contains items from every continent and period, the transformer layer is instead fixed in
advance at layer~19 (60\% depth), so no held-out continent or period influences it either. In every case the ridge
penalty is chosen within the training rows by the same leave-one-out procedure, and principal
components are fitted on training rows only. 

\paragraph{Pain directions.} The other
directions (fear, negative emotion, negative world state, bodily sensation, arousal, random,
numb, sadness) are each set's mean minus the mean of the neutral sentences, pooled over S1
and S2, and denoised against that neutral cloud. Following v2, we also rebuild every direction,
pain included, under two common baselines: against the pooled neutral sentences
(\emph{common neutral}) and against the pooled control sentences of S1 and S2
(\emph{common controls}), each denoised against its own cloud. Projections of the standalone
sets are $z$-scored against the S2 sentences and averaged over first- and third-person
versions. Removing the shared ``I feel:'' suffix changes no
value by more than $0.01$.

\paragraph{GoEmotions.} GoEmotions \citep{demszky2020goemotions} consists of Reddit comments labeled with 27 emotions or neutral, with its official train and test split (43{,}410 and
5{,}427 comments). Because comments can carry several labels, the positives for each emotion
are the comments with that label and the negatives all others. The direction is the mean of
the training positives minus the mean of the training negatives, denoised against comments
labeled only neutral, and we report the macro-average test AUC over the 27 emotions. For the
emotion-word control we remove every comment, in training and test, that contains one of the
27 label words, a WordNet synonym of one, or a derivationally related form (244 words in
all), which removes about a quarter of the comments.

\subsection{Residualizing the Transformer against the Static Baseline}
\label{sec:residual_methods}

To test whether the transformer's recoverable structure is the static structure, we fit a linear map from the 300-dimensional static embeddings $S$ to the activations $A$ at a layer, by ridge regression on the authors' training items with the penalty chosen by cross-validation on those rows, and decompose $A = SB + (A - SB)$ for all items. We call $SB$ the \emph{projection} and $A - SB$ the \emph{residual}, and probe each as we probe $A$ itself, with the same train--test split. The map is fitted on training items only, and the decomposition never sees the target. Because the projection is a linear function of $S$, it contains no linearly available information beyond $S$; it serves as a consistency check on the fit, not as a result.

Two nulls accompany the removal, because two distinct objections apply. The
\emph{variance-matched null} removes a random subspace of activation space, with its
dimension chosen so that the fraction of activation variance removed matches that of
the true removal; it answers whether a removal of this magnitude would have had this
effect regardless of what was removed. The \emph{shuffled-static null} repeats the true
removal with static vectors assigned to the wrong items; it answers whether the effect
depends on the item correspondence or is an artifact of the regression.

\subsection{Removing Group Membership from the Activations}
\label{sec:group_methods}

These analyses use the released entities, the authors' train--test split, the GloVe-covered
items and the training-selected layer of each model. Activations $A$ and static vectors $S$
are centered on training rows. Group membership $G$ is a one-hot code of country (places)
or century of death (figures); a group that occurs only in the test set gets a row of zeros.
Each removal fits a ridge map from a source matrix to a target matrix on training rows only,
with the penalty chosen on training rows by the same leave-one-out procedure, and subtracts the map's
prediction for every row, so no test item influences its own removal. \emph{Group removed}:
$R_G = A - \hat{A}(G)$, the activations minus their part predictable from group
membership. \emph{Group and static removed}: $S$ is first residualized on $G$ in the same
way, giving $S_\perp$, the static structure beyond group membership, and the part of $R_G$
predictable from $S_\perp$ is then removed. \emph{Group and random removed}: instead of the
$S_\perp$ component, a random subspace of $R_G$ (random orthonormal directions) is removed,
with as many directions as needed to remove the same share of $R_G$'s training-row variance.
For each condition a new probe is fitted on training rows and scored once on the test set;
within-group $r$ is computed after removing test-set country or century means from
predictions and targets. For the incremental analysis, $A$ and $S$ are each divided by
their mean training-row standard deviation and concatenated before probing. Intervals are
95\% percentile intervals over 1{,}000 resamples of test-set countries or centuries, with
the fitted probes held fixed. Table~\ref{tab:groupremoval} gives the full sequence.

\begin{table}[t]
  \centering
  \caption{Removing group membership from the activations (released entities, training-selected
  layers; Appendix~\ref{sec:group_methods}). $R^2$ for latitude / longitude and year of
  death, and within-country (places) or within-century (figures) $r$ of a probe refitted
  after each removal. The last row gives the within-group $r$ after removing the static
  component beyond group membership minus that after the matched random removal, with 95\%
  intervals from 1{,}000 resamples of test-set countries (273) or centuries (29).}
  \label{tab:groupremoval}
  \footnotesize
  \setlength{\tabcolsep}{4pt}
  \begin{tabular}{llcccc}
    \toprule
    & & \multicolumn{2}{c}{Places} & \multicolumn{2}{c}{Figures} \\
    \cmidrule(lr){3-4} \cmidrule(lr){5-6}
    Model & Condition & $R^2$ & Within-country $r$ & $R^2$ & Within-century $r$ \\
    \midrule
    Pythia-2.8B & Full activations             & .76 / .79 & .20 / .12 & .70 & .10 \\
                & Group removed                & .01 / .00 & .47 / .49 & .00 & .09 \\
                & Group + static removed       & .00 / .00 & .37 / .34 & .00 & .00 \\
                & Group + random removed       & .01 / .00 & .46 / .49 & .00 & .09 \\
                & Full + static appended       & .76 / .79 & .21 / .13 & .71 & .10 \\
                & Static $-$ random            & & $-.08$ [$-.12$, $-.02$] / $-.15$ [$-.19$, $-.04$] & & $-.09$ [$-.12$, $-.06$] \\
    \midrule
    Llama-2-7B  & Full activations             & .88 / .88 & .35 / .23 & .78 & .12 \\
                & Group removed                & .01 / .00 & .55 / .53 & .00 & .13 \\
                & Group + static removed       & .01 / .00 & .46 / .39 & .00 & .03 \\
                & Group + random removed       & .01 / .00 & .54 / .52 & .00 & .13 \\
                & Full + static appended       & .88 / .88 & .35 / .23 & .78 & .12 \\
                & Static $-$ random            & & $-.07$ [$-.13$, $-.03$] / $-.12$ [$-.17$, $-.04$] & & $-.10$ [$-.13$, $-.07$] \\
    \bottomrule
  \end{tabular}
\end{table}

\subsection{Grouped Holdouts on the Released Entities}
\label{sec:grouped_holdout}

For places, each item's continent is taken from its country (via the
\texttt{country\_converter} package) and the five main continents are used: Africa (2{,}406
GloVe-covered items), the Americas (9{,}953), Asia (7{,}104), Europe (9{,}614) and Oceania
(3{,}526); items whose country maps to no continent are left out. For figures, death year is
cut into five contiguous blocks of equal count: $-935$ to $1150$ ($n = 6{,}861$), $1151$--$1385$
($7{,}334$), $1386$--$1599$ ($7{,}426$), $1600$--$1814$ ($7{,}456$) and $1815$--$2021$
($7{,}507$). Each group is held out in turn, a probe is fitted on the remaining groups
(latitude and longitude jointly, as one multi-output ridge regression) and applied to the
held-out group, and $R^2$ is computed once over the pooled held-out predictions of all five
groups, each coordinate scored separately. For the transformers the layer is fixed in
advance at layer~19 (60\% depth, the depth used by \citet{gurnee2024language}) rather than
selected on the authors' training partition, which contains items from every continent and
period. $R^2$ within a single held-out group is not informative, because its denominator is
that group's small internal variance, so Table~\ref{tab:grouped_holdout} gives the median
absolute error for each held-out group. Predicting the earliest and latest blocks requires
extrapolation beyond the training range; the three middle blocks are periods surrounded by
training data, so predicting them is generalization across an unseen interval rather than
extrapolation.

\begin{table}[t]
  \centering
  \caption{Grouped holdouts on the released entities (transformers at the fixed layer~19). Median
  absolute error for each held-out group (degrees of latitude / longitude; years), the
  pooled $R^2$ over all held-out predictions, and, for comparison, $R^2$ and median error
  under random five-fold splits of the same items.}
  \label{tab:grouped_holdout}
  \footnotesize
  \setlength{\tabcolsep}{4pt}
  \begin{tabular}{lcccc}
    \toprule
    & GloVe & Word2Vec & Pythia-2.8B & Llama-2-7B \\
    \midrule
    \multicolumn{5}{l}{\emph{Places, continent held out (median error, lat.\ / long.)}} \\
    Africa        & 24.7 / 32.9 & 23.6 / 28.4 & 17.5 / 25.8 & 19.5 / 16.9 \\
    Americas      & 19.1 / 141.9 & 16.0 / 144.9 & 18.9 / 142.3 & 26.6 / 152.5 \\
    Asia          & 12.0 / 76.9 & 9.7 / 90.2 & 9.0 / 63.2 & 6.5 / 70.3 \\
    Europe        & 37.8 / 38.7 & 38.1 / 32.4 & 39.7 / 26.0 & 34.4 / 23.6 \\
    Oceania       & 61.2 / 168.2 & 59.6 / 176.1 & 45.4 / 181.0 & 54.4 / 147.8 \\
    Pooled $R^2$  & $-.52$ / $-.89$ & $-.40$ / $-.94$ & $-.31$ / $-.74$ & $-.22$ / $-.64$ \\
    Random split: $R^2$ (median error) & .48 / .55 (12 / 34) & .56 / .60 (11 / 30) & .75 / .79 (5 / 13) & .89 / .89 (3 / 8) \\
    \midrule
    \multicolumn{5}{l}{\emph{Figures, period held out (median error, years)}} \\
    $-935$--$1150$ & 644 & 647 & 538 & 509 \\
    $1151$--$1385$ & 174 & 204 & 191 & 133 \\
    $1386$--$1599$ & 156 & 138 & 137 & 108 \\
    $1600$--$1814$ & 183 & 210 & 133 & 97 \\
    $1815$--$2021$ & 398 & 413 & 328 & 337 \\
    Pooled $R^2$  & $-.11$ & $-.16$ & $.11$ & $.19$ \\
    Random split: $R^2$ (median error) & .50 (174) & .43 (191) & .68 (139) & .78 (101) \\
    \bottomrule
  \end{tabular}
\end{table}

\subsection{Items With and Without an Entity Vector}
\label{sec:coverage}

Entity vectors exist for 89\% of places and 63\% of figures, and the comparison in Table~\ref{tab:entity} uses only those items. To see whether they are easier, we trained each non-entity representation on all training items and scored it separately on held-out items with and without an entity vector (Table~\ref{tab:coverage}). Items without one are harder for every representation, the language models included, by $0.13$--$0.28$ in $R^2$; part of the difference for figures is mechanical, since the uncovered figures have a narrower range of death years (standard deviation 396 against 482 years). The drop is of similar size across representations, so the comparison between them on the common subset is not distorted, but absolute accuracy there is higher than on the full release.

\begin{table}[t]
  \centering
  \caption{$R^2$ on held-out items with and without a Wikipedia2Vec entity vector, for probes trained on all training items (authors' split, GloVe-covered items; transformers at the selected layer). Test items: 5{,}995 covered and 695 uncovered places, 4{,}710 covered and 2{,}606 uncovered figures. Counts differ slightly from Table~\ref{tab:entity}, which also requires a Wikipedia2Vec word vector.}
  \label{tab:coverage}
  \small
  \begin{tabular}{lcccc}
    \toprule
    & \multicolumn{2}{c}{Places (lat / lon)} & \multicolumn{2}{c}{Figures (year)} \\
    \cmidrule(lr){2-3} \cmidrule(lr){4-5}
    Representation & With entity vector & Without & With & Without \\
    \midrule
    GloVe, word average    & .48 / .55 & .34 / .28 & .53 & .40 \\
    fastText, word average & .59 / .62 & .43 / .34 & .61 & .47 \\
    Pythia-2.8B            & .77 / .81 & .59 / .57 & .73 & .60 \\
    Llama-2-7B             & .89 / .90 & .70 / .71 & .82 & .68 \\
    \bottomrule
  \end{tabular}
\end{table}

\subsection{Static Baseline on the Authors' Metric}

Table~\ref{tab:gt} reports the static baseline with the metric and split of the original release, for comparison with the figures \citet{gurnee2024language} report for Llama-2.

\begin{table}[!ht]
  \centering
  \caption{The type-level baseline on the entity datasets released by
  \citet{gurnee2024language}, using their designated train--test split and their
  metric (out-of-sample $R^2$; for places, the mean over the two coordinates, as in
  their evaluation code). LLM values are their reported figures for linear probes at
  60\% layer depth (their Table~2). Coverage is the share of entity names with at
  least one in-vocabulary token; each representation is scored on the test items it covers.}
  \label{tab:gt}
  \small
  \begin{tabular}{lccccc}
    \toprule
    & \multicolumn{3}{c}{World places ($n_{\text{test}} \approx 6{,}900$)} & \multicolumn{2}{c}{Historical figures ($n_{\text{test}} \approx 7{,}300$)} \\
    \cmidrule(lr){2-4} \cmidrule(lr){5-6}
    Representation & Latitude & Longitude & Both & Death year & Coverage \\
    \midrule
    GloVe            & $0.469$ & $0.526$ & $0.497$ & $0.496$ & 85\% / 98\% \\
    Word2Vec         & $0.547$ & $0.573$ & $0.560$ & $0.421$ & 88\% / 97\% \\
    fastText         & $0.586$ & $0.594$ & $0.590$ & $0.574$ & 92\% / 98\% \\
    \midrule
    Llama-2-7B       & ---     & ---     & $0.881$ & $0.785$ & --- \\
    Llama-2-13B      & ---     & ---     & $0.896$ & $0.804$ & --- \\
    Llama-2-70B      & ---     & ---     & $\mathbf{0.911}$ & $\mathbf{0.835}$ & --- \\
    \bottomrule
  \end{tabular}
\end{table}

\begin{table}[t]
  \centering
  \caption{Cosine of the S2 pain direction with other directions on static word embeddings, under the original construction and the two common baselines of v2 \citep{tagliabue2026pain}. LLM values are the means reported in v2 (23--25 models).}
  \label{tab:pain_cos}
  \small
  \begin{tabular}{lccccccc}
    \toprule
    & \multicolumn{3}{c}{S2 $\times$ fear} & \multicolumn{3}{c}{S2 $\times$ negative emotion} & S2 $\times$ sadness \\
    \cmidrule(lr){2-4} \cmidrule(lr){5-7} \cmidrule(lr){8-8}
    Representation & Original & Neutral & Controls & Original & Neutral & Controls & Original \\
    \midrule
    GloVe    & $+0.05$ & $+0.61$ & $+0.11$ & $+0.23$ & $+0.76$ & $+0.08$ & $+0.19$ \\
    Word2Vec & $+0.03$ & $+0.54$ & $+0.13$ & $+0.19$ & $+0.71$ & $+0.10$ & $+0.10$ \\
    fastText & $-0.01$ & $+0.61$ & $+0.03$ & $+0.14$ & $+0.71$ & $+0.06$ & $+0.11$ \\
    \midrule
    LLMs (mean) & $+0.12$ & $+0.58$ & $+0.08$ & $+0.21$ & $+0.77$ & $+0.14$ & $+0.38$ \\
    \bottomrule
  \end{tabular}
\end{table}

\begin{figure}[t]
  \centering
  \includegraphics[width=\textwidth]{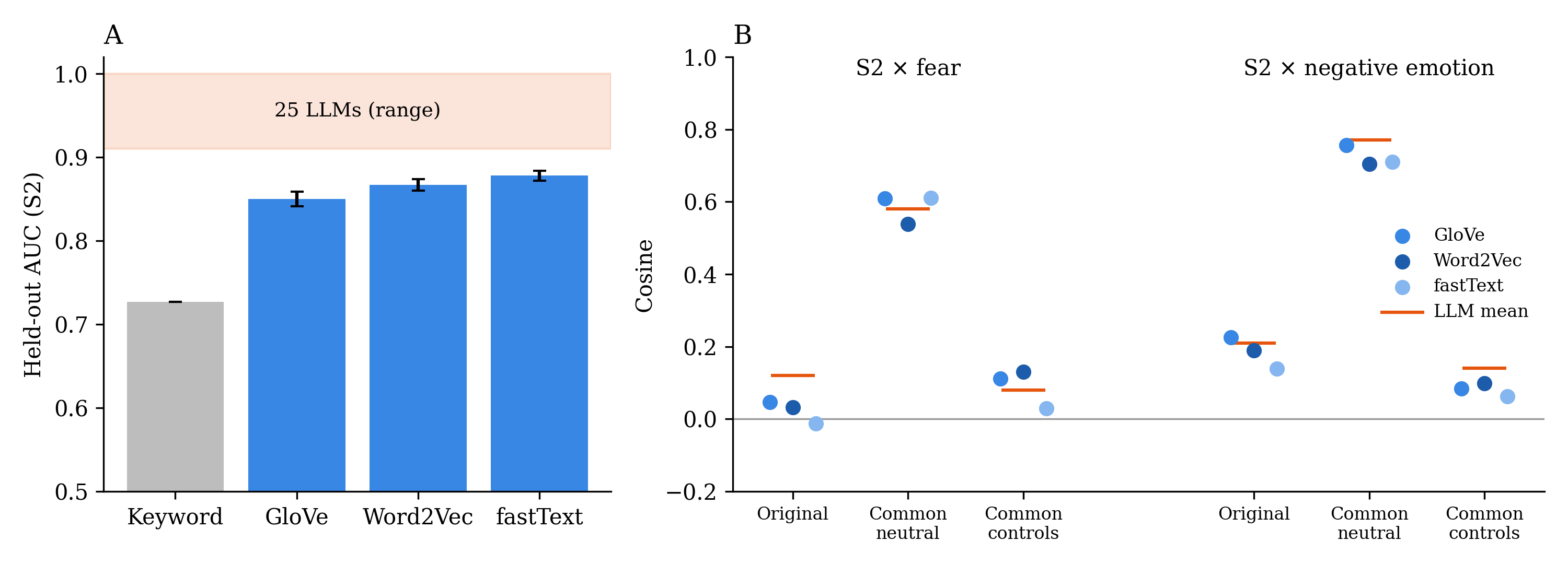}
  \caption{The representational tests reported for a ``pain axis'' in 25 LLMs
  \citep{tagliabue2026pain}, run on averaged static word vectors of the authors' own
  sentences. \emph{A}: held-out separation of pain from controls (S2, first person;
  five-fold cross-validation over sentence sets, mean $\pm$ s.d.\ over 20 fold
  assignments). The keyword baseline is the best of several surface-lexical classifiers
  reported by an independent review \citep{painreview2026}. LLM values are each model's
  best layer. \emph{B}: cosine of the pain direction with fear, negative-emotion and
  negative-world-state directions; bars span the LLMs' per-model range.}
  \label{fig:pain}
\end{figure}

\subsection{Group Structure of the Historical Figures}
\label{sec:figure_groups}

On the released figures, year of death varies almost entirely between centuries
($\eta^2 = 0.996$ in the test set) and mostly between three eras (before 500\,CE,
500--1400, after 1400; $\eta^2 = 0.80$). As oracle decompositions derived from the target itself, an
era label accounts for $R^2 = 0.80$ and a century label for $1.00$, against $0.50$ for GloVe. Replacing each figure's GloVe vector
with its era's training centroid gives $0.80$, and the within-era residual gives $0.04$.
Within centuries, GloVe and Word2Vec order figures at $r = 0.05$.

\subsection{Semantic Ablation on the Released Places}
\label{sec:semantic_ablation}

To ask which kinds of words the coordinate probe depends on, we projected out the embedding subspace spanned by a word category and refit the probe. For each category we took the GloVe vectors of its words, kept the principal components covering 90\% of their variance (at most 20), and projected that subspace out of every place's averaged name vector; the place names themselves are unchanged. We trained a ridge probe on the training places (authors' split) and scored $R^2$ on the test places ($n = 33{,}500$ places with a GloVe vector; baseline $R^2 = 0.47$ for latitude and $0.53$ for longitude). The null is 200 random word sets matched to the category on number of words, frequency decile, whether the capitalized form is the more frequent one, and vector norm, each reduced to the same number of dimensions. Projecting out the country-name subspace costs the most, about three times the matched null for latitude, and the region and continent and the climate subspaces cost more than the null for latitude. Cardinal directions, economic terms and cultural and language words cost no more than matched random words. The drops are nevertheless small in absolute terms: projecting out roughly 20 of 300 dimensions leaves most of the decodable coordinate signal intact, so the signal is not confined to these semantic subspaces.

\begin{table}[!ht]
  \centering
  \caption{Drop in test $R^2$ when the embedding subspace spanned by a word category is projected out of the GloVe name vectors of the released places, with the mean drop for 200 matched random word sets in parentheses and $z$ against that null. $k$ is the number of dimensions projected out and Var.\ the share of the place vectors' variance they hold. $^{*}$\,$p \le .05$ and $^{**}$\,$p \le .01$, one-sided empirical $p$ over the 200 matched sets.}
  \label{tab:semantic_ablation}
  \small
  \setlength{\tabcolsep}{4pt}
  \begin{tabular}{lcccccccc}
    \toprule
    & & & & \multicolumn{2}{c}{Latitude} & \multicolumn{2}{c}{Longitude} \\
    \cmidrule(lr){5-6} \cmidrule(lr){7-8}
    Category & Words & $k$ & Var. & Drop (null) & $z$ & Drop (null) & $z$ \\
    \midrule
    Cardinal directions & 16 & 9 & 0.07 & .013 (.007) & 1.3 & .005 (.013) & $-$1.0 \\
    Climate and weather & 27 & 19 & 0.11 & .019 (.008)$^{*}$ & 3.0 & .004 (.008) & $-$1.2 \\
    Region and continent & 28 & 18 & 0.14 & .053 (.022)$^{*}$ & 2.6 & .037 (.032) & 0.3 \\
    Country names & 68 & 20 & 0.17 & .095 (.032)$^{**}$ & 5.9 & .091 (.055)$^{*}$ & 2.6 \\
    Economic terms & 27 & 19 & 0.11 & .008 (.008) & 0.0 & .006 (.009) & $-$0.5 \\
    Cultural and language & 31 & 20 & 0.15 & .019 (.020) & $-$0.2 & .017 (.029) & $-$1.0 \\
    \bottomrule
  \end{tabular}
\end{table}

\end{document}